\documentclass[sigplan,screen]{acmart}
\usepackage{adjustbox}
\usepackage{caption}
\usepackage{subcaption}
\usepackage{colortbl}
\usepackage{graphicx}
\usepackage{amsmath}
\usepackage{algorithm2e}
\usepackage{blindtext}
\usepackage{tcolorbox}
\usepackage{multirow}
\usepackage{hhline}
\usepackage{xcolor, soul}
\usepackage{graphicx}
\usepackage[font={large,it}]{caption}
\makeatletter
\def\BState{\State\hskip-\ALG@thistlm}
\makeatother
\usepackage[font=small,skip=10pt]{caption}
\usepackage{array}
\newcolumntype{P}[1]{>{\centering\arraybackslash}p{#1}}
\AtBeginDocument{%
  \providecommand\BibTeX{{%
    \normalfont B\kern-0.5em{\scshape i\kern-0.25em b}\kern-0.8em\TeX}}}

\setcopyright{none} 

\begin{document}
\settopmatter{printacmref=false}

\title{Developing a novel fair-loan-predictor through a multi-sensitive  debiasing pipeline: DualFair}

\author{Jashandeep Singh}
\email{jashan@mit.edu}
\orcid{0000-0001-8186-7383}
\authornotemark[1]
\affiliation{%
  \institution{Floyd B. Buchanan High School} 
  \streetaddress{77 Massachusetts Ave}
  \city{Fresno}
  \country{USA}
  \postcode{02139}
}

\author{Arashdeep Singh}
\email{asingh@csail.mit.edu}
\orcid{0000-0002-6109-451X}
\affiliation{%
  \institution{Floyd B. Buchanan High School} 
  \streetaddress{77 Massachusetts Ave}
  \city{Fresno}
  \country{USA}
  \postcode{02139}
}

\author{Ariba Khan}
\email{akhan02@mit.edu}
\orcid{0000-0002-7613-5636}
\affiliation{
  \institution{Massachusetts Institute of Technology} 
  Computer Science and Artificial Intelligence Laboratory 
  \streetaddress{77 Massachusetts Ave}
  \city{Cambridge}
  \country{USA}
  \postcode{02139}
}

\author{Amar Gupta}
\email{agupta@mit.edu}
\orcid{0000-0001-9306-1256}
\affiliation{
  \institution{Massachusetts Institute of Technology} 
  Computer Science and Artificial Intelligence Laboratory 
  \streetaddress{77 Massachusetts Ave}
  \city{Cambridge}
  \country{USA}
  \postcode{02139}
}

\begin{abstract}
Machine learning (ML) models are increasingly used for high-stake applications that can greatly impact people’s lives. Despite their use, these models have the potential to be biased towards certain social groups on the basis of race, gender, or ethnicity. Many prior works have attempted to mitigate this “model discrimination” by updating the training data (pre-processing), altering the model learning process (in-processing), or manipulating model output (post-processing). However, these works have not yet been extended to the realm of multi-sensitive parameters and sensitive options (MSPSO), where sensitive parameters are attributes that can be discriminated against (e.g race) and sensitive options are options within sensitive parameters (e.g black or white), thus giving them limited real-world usability. Prior work in fairness has also suffered from an accuracy-fairness tradeoff that prevents both the accuracy and fairness from being high. Moreover, previous literature has failed to provide holistic fairness metrics that work with MSPSO. In this paper, we solve all three of these problems by \textbf{(a)} creating a novel bias mitigation technique called DualFair and \textbf{(b)} developing a new fairness metric (i.e. AWI) that can handle MSPSO. Lastly, we test our novel mitigation method using a comprehensive U.S mortgage lending dataset and show that our classifier, or fair loan predictor, obtains better fairness and accuracy metrics than current state-of-the-art models. 
\end{abstract}

 \maketitle


\section{Introduction}
Machine learning (ML) models have enabled automated decision-making in a variety of fields, ranging from lending to hiring to criminal justice. However, the data often used to train these ML models contain many societal biases. These biased models have the potential to perpetuate stereotypes and promote discriminatory practices, therefore giving privileged groups undue advantages. As a result, it has become increasingly important for ML researchers and engineers to work together in eliminating this algorithmic unfairness. 

Despite an awareness of the need for these fair models, there still exists many examples of models exhibiting prevalent biases. For example: 

\begin{itemize}

\item
In 2016, ProPublica reported that ML models used by judges to decide whether to keep criminals in jail were discriminatory against African-Americans males, labeling them with relatively high recidivism scores \cite{angwinMachineBiasProPublica2016}.

\item
Amazon discovered  its automated recruiting system was biased against female job applicants, rendering them far less successful in the application process \cite{dastinAmazonScrapsSecret2018}.

\item
A healthcare algorithm evaluated on 200 million individuals to predict whether patients needed extra medical care was highly discriminatory against African Americans while prioritizing white individuals \cite{ledfordMillionsBlackPeople2019}.

\end{itemize}

One particular domain where bias mitigation has become especially crucial is mortgage lending. It has been reported that over 3.2 million mortgage loan applications and 10 million refinance applications exhibited bias against Latinx and African American lenders \cite{weberBlackLoansMatter}. In another study, it was shown that minority groups were charged significantly higher interest rates (by 7.9 basis points) and were rejected 13\% more often than their privileged counterparts. These biases, when trained upon, lead to discriminatory loan predictors that emphasize the gap in suitable housing, wealth, and property between unprivileged (African Americans) and privileged (White and Asian American) groups. This problem becomes even more exacerbated as companies widely begin to use these biased models in real-life decisions. Additionally, these biased models are unlawful under the Equal Credit Opportunity Act (ECOA), which forbids the discrimination of individuals based upon sensitive attributes (e.g race, gender, national origin, and ethnicity) by any private or public institution. It has thus become the moral and legal duty for researchers and software developers to find a solution to this problem of “algorithmic unfairness.” 

Fortunately, work has been done in tackling ways to mitigate this bias. There are currently three ways bias mitigation has been approached, correlating with before, in, and after the data usage pipeline:

\begin{itemize}
\item
\emph{Pre-processing} — the transformation of data (e.g the alteration of sensitive attribute sampling distributions) to “repair” inherent biases \cite{kusnerCounterfactualFairness2018,chiappaPathSpecificCounterfactualFairness2018, brunetUnderstandingOriginsBias2019, calmonOptimizedPreProcessingDiscrimination2017}

\item
\emph{In-processing} — the use of classifier optimization or fairness regularization terms to affect model learning and maximize a model’s fairness \cite{zhangMitigatingUnwantedBiases2018, wuFairnessawareClassificationCriterion2018, berkConvexFrameworkFair2017}. 

\item 
\emph{Post-processing} — the manipulation of model output to improve performance and fairness metrics \cite{hardtEqualityOpportunitySupervised2016}. 
\end{itemize}

While many prior works gain relatively high fairness and performance metrics using one or more of these mitigation techniques, the current literature contains three main problems that hamper their adaptability and deployment: (1) little work has been done on extending mitigation techniques to situations with \emph{multi-sensitive parameters and sensitive options (MSPSO)}, (2) an accuracy and fairness tradeoff still exists, and (3) an absence of accepted fairness metrics for MSPSO data. 

\textbf{Our Contributions}: In this paper, we target all four of the previously stated problems to develop a novel and real-world applicable, fair ML classifier in the mortgage lending domain that obtains state-of-the-art fairness and accuracy metrics. Through this process, we coin a bias-mitigation pipeline called DualFair (a pre-processing strategy), which approaches fairness through a new lens, and  solves many problems hindering the growth of the "Fairness, Accountability, and Transparency", or FAT, field.

More concretely, the main insights we provide within this paper includes the following: 

\begin{itemize}
\item
Creating a bias mitigation strategy coined DualFair, which debiases data through oversampling and undersampling techniques that target root causes of bias. 
\item
Extending our mitigation approach to MSPSO (multi-sensitive parameters and sensitive options) by subdividing datasets and then balancing by class and labels. 
\item 
Developing a novel fairness metric called AWI (Alternate World Index) that is devoid of bias, generalizable to MSPSO, and an accurate representation of model fairness.
\item
Eliminating the \emph{accuracy-fairness trade-off} by debiasing our mortgage lending data using DualFair. 
\item
Creating a fair loan predictor that achieves state-of-the-art fairness and accuracy metrics that can be used by practitioners.  
\end{itemize}

The rest of this paper is structured as follows: Section 2 provides an overview of prior work and contributions in the FAT field, particularly in relation to our own. Section 3 explains fairness terminology and metrics used throughout our paper. Section 4 gives a detailed outline of our bias mitigation approach, DualFair, and our novel fairness metric AWI. Section 5 summarizes the results of our bias mitigation pipeline and documents the success of our approach compared to the previous state-of-the-art based upon \emph{accuracy, precision, false alarm rate, recall, F1 score and AWI}. In Section 6, we give a brief overview on potential directions for future work. Finally, Section 7 concludes the paper.

\section{Related Work}
Fairness in ML models is a prevalently explored topic within the AI community. Recently, major industries have begun to put a greater priority on AI fairness. IBM developed AI Fairness 360, which is a fairness toolkit that contains commonly used fairness metrics and debiasing techniques to aid researchers working in ML fairness \cite{bellamyAIFairness3602018}. Microsoft has established FATE \cite{MicrosoftTrustFATE}, a research group dedicated to fairness, accountability, transparency, and ethics in AI. Google \cite{ResponsibleAIPutting2019}, Microsoft \cite{madaioCoDesigningChecklistsUnderstand2020}, IEEE \cite{chatilaIEEEGlobalInitiative2019}, and The European Union \cite{weiserBuildingTrustHumancentric2019} each respectively published on ethical principles in AI, which are general guidelines on what the companies define as “responsible AI.” Facebook has created bias detection tools for their own internal AI systems \cite{gershgornFacebookSaysIt}. The research community has started to take an interest in fair AI as well. ASE 2019 and ICSE 2018 have hosted workshops on software fairness \cite{EXPLAIN2019ASE}. Mehrabi et al. have studied various notions of fairness and fundamental causes of bias \cite{mehrabiSurveyBiasFairness2021}. ACM has established the FAccT ‘21 as a conference to spearhead work on fairness and transparency of ML algorithms \cite{ACMFAccT}. 

Thus far, achieving algorithmic fairness has been addressed through pre-processing, in-processing, and post-processing approaches. Prior work has proposed a variety of bias mitigating methods implementing these approaches. \emph{Optimized pre-processing} \cite{calmonOptimizedPreProcessingDiscrimination2017} is a pre-processing method that seeks to achieve group fairness by altering labels and features using probabilistic transformations. Zhang et al. presented the in-processing approach \emph{Adversarial Debiasing} \cite{kennaUsingAdversarialDebiasing2021}, which increases accuracy and strips away an adversary’s ability to make decisions based upon protected attributes using GANs. \emph{Reject option classification} \cite{kamiranExploitingRejectOption2018} is a post-processing strategy that translates favorable outcomes from the privileged group to the unprivileged group and unfavorable outcomes from the unprivileged group to the privileged group based upon a certain level of confidence and uncertainty. Chakraborty et al. proposed \emph{Fair-SMOTE} \cite{chakrabortyBiasMachineLearning2021}, a pre-processing and in-processing approach, which balances class and label distributions and performs \emph{situation testing} (i.e testing individual fairness through alternate “worlds”). 

Our experience has shown, however, these approaches lack mainly in their inability to extend to MSPSO. Chakraborty et al. noted that the consideration of MSPSO would divide data into unmanageable small regions \cite{chakrabortyFairwayWayBuild2020}. Salerio et al. attempted to approach MSPSO by creating AEQUITIS, a fairness toolkit, that uses parameter-specific fairness metrics to systematically view bias within one sensitive parameter at a time \cite{saleiroAequitasBiasFairness2019}.  Gill et al., Chakraborty et al., and Kusner et al. approached fairness in their own separate domains by using a singular sensitive parameter, designating one privileged group and one unprivileged group as a way to compare mitigation results \cite{kusnerCounterfactualFairness2018}. Through our experience with DualFair, we argue that it is possible to debias MSPSO data given a proper pipeline, approach, and data. This allows for deployability and scalability within real-world systems. We also show that given MSPSO data, one could devise a fairness metric (AWI), which considers bias from all parameters and options cohesively. 

The mortgage domain has seen its own work in the realm of AI fairness as well. Fuster et al. and Bartlet et al. showed an immense disparity in over 92\% of loans, spanning originations, interest rate charges, and preapprovals across the United States on the basis of sex, race, and ethnicity \cite{fusterPredictablyUnequalEffects2021, bartlettConsumerLendingDiscriminationFinTech}. Gill et al. built upon these conclusions and proposed a state-of-the-art machine learning workflow that can mitigate discriminatory risk in singular sensitive parameter and sensitive option mortgage data while maintaining interpretability \cite{gillResponsibleMachineLearning2020}. This framework was used to create a fair-loan-predictor. Lee et al. also presented a theoretical discussion of mortgage domain fairness through relational tradeoff optimization \cite{leeAlgorithmicFairnessMortgage2021}. That is, the paper discussed a method to achieve a balance between accuracy and fairness within the \emph{accuracy-fairness tradeoff} on mortgage lending data rather than maximizing both. 

Our work builds on the foundation created by these previous works in mortgage lending and bias mitigation in AI systems at large. It is important to note that the literature varies in two main aspects, however:

\begin{itemize}
    \item 
     finding bias 
    \item
     mitigating bias
\end{itemize}

While most prior work have centralized on finding bias, our study seeks to mitigate bias through creating a debiasing pipeline and training a fair-loan-predictor. 

\captionsetup[table]{font=large, textfont={bf}}
\captionsetup[figure]{font=large, textfont={bf}}

\begin{table*}
\caption{Population statistics for HMDA (note: all values are expressed in \$1000, Interest Rate is expressed as a percentage, and LTV is expressed as a ratio; lastly, groups are split via sensitive parameters sex, race, and ethnicity respectively)}
\begin{adjustbox}{width= 14cm, height = 3.0cm}
\centering
\begin{tabular}{lclccccc} 
\hline
Group      &  &        & Income & Loan Amt. & Interest Rate & LTV Ratio & Property Value~  \\ 
\hline
Male       &  & Mean   & 124    & 267       & 3.36\%          & 76.4                & 405              \\
(n = 1,224,719)  &  & Median & 84     & 225       & 3.19\%            & 79.8                & 315              \\ 
\hline
Female     &  & Mean   & 89     & 225       & 3.68\%            & 74.4                & 346              \\
(n = 805,583)  &  & Median & 69     & 195       & 3.25\%            & 78.6                & 275              \\ 
\hline
White      &  & Mean   & 131    & 270       & 3.41\%            & 73.6                & 428              \\
(n = 3,474,562)  &  & Median & 96     & 235       & 3.13\%            & 76.1                & 335              \\ 
\hline
Black      &  & Mean   & 94     & 240       & 3.47\%            & 83.5                & 319              \\
(n = 220,517)  &  & Median & 76     & 215       & 3.25\%            & 90                  & 265              \\ 
\hline
Non. Hisp. &  & Mean   & 133    & 271       & 3.42\%            & 73.6                & 430              \\
(n = 3,360,377)  &  & Median & 97     & 235       & 3.13\%            & 76.2                & 335              \\ 
\hline
Hispanic   &  & Mean   & 96     & 256       & 3.36\%            & 79.3                & 362              \\
(n = 334,184)  &  & Median & 76     & 235       & 3.25\%            & 80.0                & 305              \\
\hline
\end{tabular}
\end{adjustbox}
\end{table*}

\section{Fairness Terminology}
In this section, we outline fairness terminology that will be used within this work. An \emph{unprivileged group} is one that is discriminated against by a ML model. \emph{Privileged groups} are favored by a ML model due to some sensitive parameter. These groups usually receive the \emph{favorable label} (i.e the label wanted), which, for our purposes, is a mortgage loan application being accepted. A \emph{sensitive parameter}, also known as a \emph{protected attribute}, is a feature that distinguishes a population of people into two groups, an unprivileged and privileged group. This parameter usually has been historically discriminated against (e.g race and sex). \emph{Sensitive options} are sub-groups, or options, within sensitive parameters (e.g for race: White, Black, or Asian). The distribution of all sensitive parameters and sensitive options (e.g White Males, Black Males, White Females, Black Females) is referred to as a \emph{class distribution}. The difference in favorable outcomes and unfavorable outcomes for a particular group as represented by the ground-truth is its \emph{label balance}. \emph{Label bias} is a type of societal bias which can shift the label balance (e.g, a mortgage underwrite subconsciously denying a reliable African-American lenders for a loan). \emph{Selection bias} is another type of bias that is created when selecting a sample. For example, suppose one is collecting car insurance data for a particular location. However, the particular location they are collecting data from has some historical discrimination that causes it to have a low annual income per person. The insurance data collected in this location, therefore, would contain implicit economic biases. \emph{Fairness metrics} are a quantitative definition of bias within a specific dataset. 

Finally, there are two main types of fairness: \emph{individual fairness} and \emph{group fairness} \cite{mehrabiSurveyBiasFairness2021}. 
\begin{itemize}
    \item 
     Individual fairness is when similar individuals obtain similar results. 
    \item
    Group fairness is when the unprivileged and privileged group, based upon a particular sensitive parameter, will be treated similarly. 
\end{itemize}

Before beginning our discussion on DualFair, we would like to note that in this work we use a binary classification model for all of our inferences and methods of achieving these notions of fairness. Future work could make an effort of looking into algorithmic fairness with regression models instead.

\section{Materials and Methods}

\subsection{Mortgage Data}

Previous domain-specific ML studies have faced many challenges in acquiring large-scale datasets for comprehensive work. The mortgage domain conveniently offers a solution to this problem. For our study, we use the HMDA dataset, which was publicly made available by the Home Mortgage Disclosure Act (HMDA) of 1975 \cite{HMDADataBrowser}. HMDA data has been used in various studies to outline and understand recent mortgage trends. It spans 90\% of all loan originations and rejections in the United States and contains over 21 distinct features (e.g., race, sex, ethnicity).

It is important to note that HMDA is not entirely devoid of bias and has racial prejudices prevalent within the data. Table 1 shows a quantitative distribution of these biases through particular sensitive, or biased, features. Our evaluation shows that features such as loan amount, income, and interest already give certain groups an undue advantage, which will be dealt downstream during DualFair’s bias mitigation pipeline. 

In this study, we use HMDA national data from 2018 to 2020 and 2020 data from 2 small states (<150,000 rows of data), 2 medium states (>150,000 to <250,000 rows of data), and 2 large states (>250,000 rows of data), totaling over 22 million home loan applications. This data is unique compared to past years’ as in 2018 the Dodd-Frank Wall Street Reform and Consumer Protection Act (Dodd-Frank) mandated more expansive updates to mortgage loan data from all applicable institutions. Dodd-Frank led to the addition of features such as credit score, debt-to-income ratio, interest rate, and loan term, providing a more comprehensive review of loan applicants and studies for algorithmic fairness. In our work, we build a fair ML classifier on the HMDA data using DualFair, where our classifier predicts whether an individual originated (i.e y = 1) or was denied (i.e y = 0) a loan. The following steps were taken to facilitate the creation of this classifier and analysis of HMDA data: 

\begin{itemize}
\item
5 million loan records were randomly sampled without replacement from each year 2018-2020 to form a combined HMDA dataset, spanning over 15 million loan applications, for analysis. 
\item
Features with more than 25\% data missing, exempt, or not available were removed during data pre-processing. 

\item 
Only White, Black, and Joint labels from the race category, Male, Female, and Joint labels from the sex category, and Non-Hispanic or Latino, Hispanic or Latino, and Joint labels from the ethnicity category were used in the study.
\end{itemize}
Note: Joint is defined as a co-applicant sharing a different feature option than the central applicant. For instance, a White male applicant and a Black co-applicant would be joint for race. Future research is highly encouraged in implementing all sensitive parameters and sensitive options, where our code would be able to generalize.

\subsection{Debiasing}
It has been shown that data bias mainly derives from two factors: \emph{label bias} and \emph{selection bias}. In this section, we eliminate both of these biases from our dataset in the debiasing process through a novel pre-processing approach we coin DualFair. More concretely, DualFair will remove the \emph{accuracy-fairness tradeoff} and increase algorithmic fairness metrics by approaching the task of selection bias and label bias through balancing at the class and label level. 

\begin{figure*}[t]
  \includegraphics[width=8.77cm, height = 8cm]
    {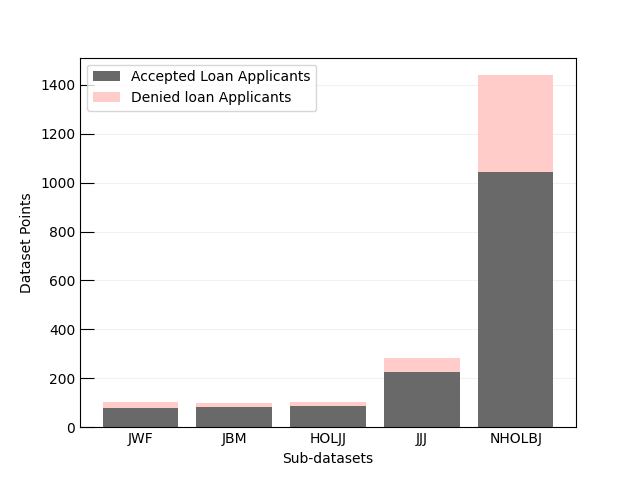}
    \includegraphics[width=8.77cm, height = 8cm]
    {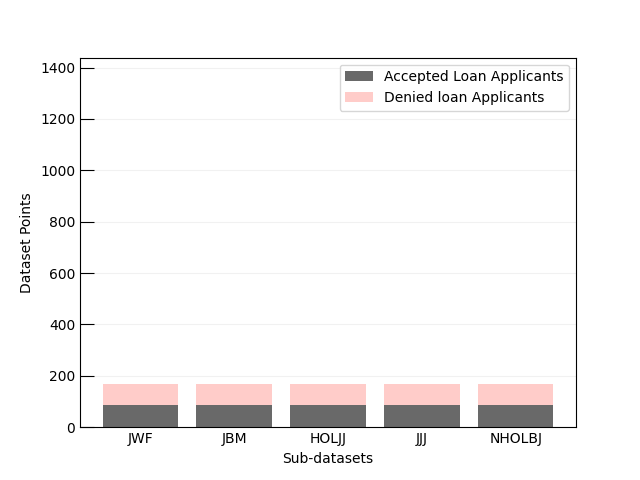}

  \caption{Before, After Class, and Label Balancing of DualFair (note: only 5 subdatasets from the balancing procedure of Connecticut, a smaller sample of HMDA, are shown lui the normal 27; also, note that J denotes Joint, B denotes Black, W denotes White, F denotes Female, M denotes Male, and HOL denotes Hispanic or Latino )}
\end{figure*}

At first, DualFair will split the central dataset into sub-datasets based upon all combinations of sensitive parameters and sensitive options. The following methodology is used to designate sub-datasets: suppose a dataset contains two sensitive parameters, sex and race. Also suppose there are only two sensitive options for each sensitive parameter, \emph{Male (M)} or \emph{Female (F)} and \emph{Black (B)} or \emph{White (W)}. After being split, the dataset, D, would be broken into four distinct sub-datasets \emph{WM (White Males)}, \emph{BM (Black Males)}, \emph{WF (White Females)}, \emph{BF (Black Females)}. 

In the case of HMDA, DualFair would result in 27 sub-datasets. We start with 3 sensitive parameters, race, sex, and ethnicity.  Then upon each parameter, there is a division by sensitive options, where each sensitive parameter has three different options. For race, an individual could be White, Black, or Joint; for sex, individuals could be Male, Female, or Joint; and, lastly, for ethnicities, individuals could be Non-Hispanic or Latino, Hispanic or Latino, or Joint. 

We can generate an equation to represent the count of sub-datasets given a vector of sensitive parameters, $p$, and a vector of sensitive options, $o$: 

\[ \hspace*{0.1cm} 
p = \left[\begin{array}{cc}
    p^1\\
    p^2\\
    p^n\\
\end{array}\right]
\hspace*{0.2cm}
o = \left[\begin{array}{cc}
    o^1\\
    o^2\\
    o^n\\
\end{array}\right] \]
\vspace{.25 cm}

\hspace*{2.3cm} $p \cdot o$ = $sub\-datasets$ $count$ \hspace{1 cm} (1)

\vspace{.25 cm}

In our case, $p$ is $[1, 1, 1]$ and $o$ is $[3, 3, 3]$; therefore, vector multiplication (i.e $p \cdot o$) gives us 27 sub-datasets. 

It is worthwhile to note that DualFair’s algorithm is highly efficient at generating sub-datasets obtaining a $O(n^2)$ in terms of computational processing; hence, it is very much scalable to MSPSO datasets spanning many sensitive parameters and options.

After, in each subset of data, we obtain the number of accepted ($y = 1$) and rejected ($y = 0$) labels. We take the median number of accepted and rejected labels in all subsets of data and synthetically oversample or undersample each class in the sub-datasets to that value using SMOTE. The result of this process as shown in Figure 1 is a class-balanced HMDA dataset that has no selection bias. 

Figure 2 captures the class distributions before and after selection bias were removed. Observing the figure, it can be seen that HMDA is unbalanced within the class distributions (i.e between different subsets of data) and label distribution (i.e balance of accepted and rejected labels within a class). Some sub-datasets are far more common in the data while some are not. Thus, there is a huge imbalance and presentation of root biases that need to be taken into account. In the after DualFair snapshot of the figure, it is shown that all sub-datasets have become balanced at both the label and class distribution level. This strips away selection bias by regulating the way the model perceives the data (i.e making it so no data overly or underlie trained upon). 

In using oversampling techniques, we follow the general guidelines taken strictly from Chakraborty et al. Thus, we preserve valuable associations between values when oversampling. When creating data, we make sure it is close to neighboring examples. This allows for the average case association that may be between two particular variables. We use two hyperparameters called “mutation amount” ($f$) and “crossover frequency” ($cr$) to carefully use SMOTE. These parameters lie in between [0, 1]. “Mutation amount” controls the probability the new data point is different from the parent point while “crossover frequency” denotes the probability of how different the new data point is to its paternal point. We utilize .8 (80\%) as the value for both of these parameters as it serves the best in terms of preserving vital associations or results. 

\begin{figure}[t]
  \centering
  \includegraphics[width= 7cm, height = 11.5cm]{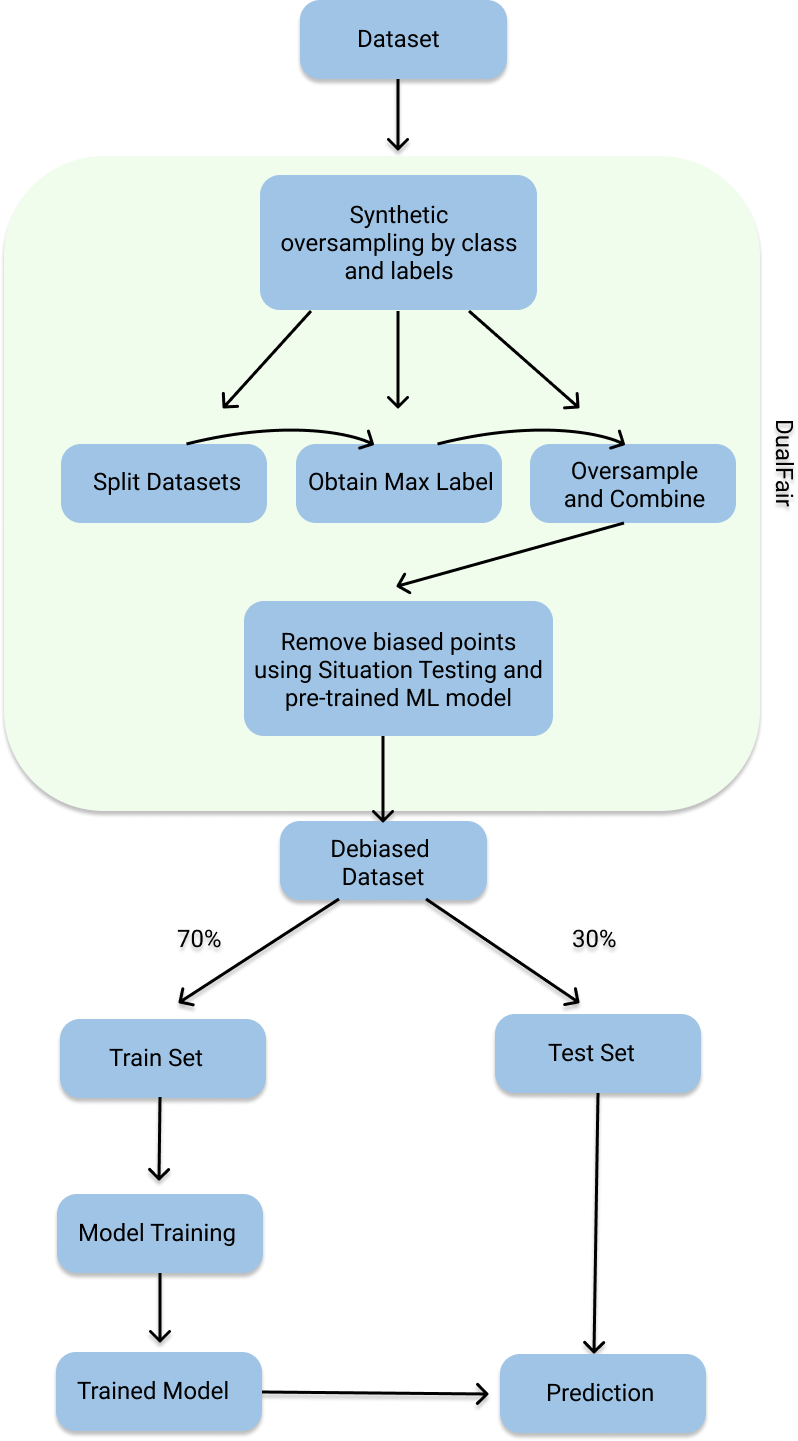} 
  \caption{
DualFair Bias Mitigation Pipeline  
}
\end{figure}

After balancing, our debiasing process uses a method known as situation testing coined in Chakrabokty et al. to reduce label bias. Situation testing finds biased points within the training data by probing all combinations of sensitive options within the data. More clearly, situation testing will test all combinations of sensitive options on an ML model trained on the balanced dataset. If the model predicts a unique value in any of the theoretical counterfactual “worlds,” then the point is removed. This process removes biased data points from the dataset and decreases label bias.

The use of balancing and situation testing is described by the DualFair process outlined in Figure 1. After removing label and selection bias from our dataset, we will have a debiased dataset that can be propagated into our framework for yielding metrics. Through DualFair, this dataset has no accuracy-fairness tradeoff. More specifically, the accuracy-fairness tradeoff is eliminated as DualFair creates fair data to evaluate and train from. Fairness metrics will be shown for the DualFair process in section 3.4.

\subsection{Novel Fairness Metrics}

Although there are a significant amount of fairness metrics proposed by the literature, all of the fairness metrics lack in one general area: non-applicability to MSPSO data. We remedy this issue by creating a new fairness metric, the Alternate World Index (AWI), based upon computational truths and previous literature. 

Our research has also shown that standard fairness metrics such as Equality of Opportunity and Equalized Odd fail to capture \emph{individual} and \emph{group fairness} by disregarding denied applicants. However, since this is not a focal point of our work, we will leave the discussion here for future work to elaborate upon. 

Let us begin by defining fairness. In prior work, there has been a lack of a universal definition of fairness, so for our purposes, we will define fairness as having different sensitive groups (e.g male group and female group) being treated equivalently and similar individuals (i.e possess similar statistics) be treated equivalently. For this definition of fairness to occur, both \emph{group fairness} and \emph{individual fairness} must be met.

\begin{itemize}
\item
Group fairness is the goal that based on sensitive parameters, unprivileged and privileged groups will be treated similarly. 
\item
Individual fairness means that similar individuals will have similar outcomes
\end{itemize}

Mathematically, this fairness is defined as {$U_g | y = 1$} being similar to  {$P_g | y = 1$} where $U_g$ is the underprivileged groups, $P_g$, is the privileged groups, and $y = 1$, represents the desired outcome. 

Using our previous definition of fairness, we can generate a metric that satisfies the interpretation of the previous requirements for MSPSO work; we coin this metric AWI (Alternate World Index).

\[ \hspace*{0.4cm} \sum_{n=1}^{n} \sum_{n=1}^{27} \frac{p_k}{27} \hspace*{0.2cm} \epsilon \hspace*{0.2cm} \{0, 1\} \hspace*{0.5cm} (2) \]

\vspace{.135 cm}

\[ \hspace*{0.4cm} n = number\ of\ data\ points \hspace*{0.5cm}  \]

Specifically, AWI is a count of the number of biased points within a dataset normalized by the dataset size. A biased point is identified by iterating a point through all respective counterfactual worlds, in our case 27, and evaluating any disparity in model prediction. A simple visualization of this process for bias point classification is shown in Figure 5. If under all counterfactuals, model prediction is constant we call this point devoid of bias; however, if one situation yields a unique result the point is marked as ambiguous (biased). After repeating this process for all points in the test dataset, we use the count of total bias points and total dataset points to yield AWI. In our study, we report AWI 10 times larger than its value to more accurately represent its difference, whether beneficial or harmful.

AWI extends fairness metrics to the realm of MSPSO by quantifying the biased points within a dataset using counterfactuals. By doing this, we solve two major problems when applying fairness metrics to MSPSO data: (1) an unequal amount of privileged to unprivileged group and (2) the lack of one holistic fairness value for an entire dataset.

AWI takes a pragmatic approach to fairness evaluation, specifically targeting similar individuals and their different realities. It is a versatile metric for any work looking to achieve individual or group fairness with or without MSPSO. One pitfall of AWI is its computational expense, especially with large volumes of data, because it runs 27 iterations per row of data. Future research could look for directions in optimization.

\RestyleAlgo{ruled}

\begin{algorithm}[hbt!]
\caption{Pseudocode for Alternate World Index}\label{alg:two}
\textbf{Input:} df, globaldf \\
\textbf{Output:} AWI\\

\SetAlgoLined
\SetKwProg{Def}{def}{}{end}
\Def{get\_AWI ($df$, $globaldf$)}{

    $df_{train}, df_{test} \gets df$\\ 
    $X_{train}, Y_{train} \gets df_{train}$\\
    $clf \gets X_{train}, Y_{train}$\\
    $removal\_list = []$\\
    \For{index, row in df\_test}{
        $pred\_list = []$\\
        $row_ = all\_features\_except\_label$\\
        \For{global\_index, global\_row in global\_df}{
            $current\_comb  = get\_entire\_row$\\
            $orig\_ethnic = row\_[0][2]$\\
            $orig\_race = row\_[0][3]$\\
            $orig\_sex = row\_[0][4]$\\
            $row\_[0][2] = current\_comb[0]$\\
            $row\_[0][3] = current\_comb[1]$\\
            $row\_[0][4] = current\_comb[2]$\\
            $y\_current\_pred = clf.predict(row\_)[0]$\\
           $ pred\_list.append(y\_current\_pred)$\\
            $row\_[0][2] = orig\_ethnic$\\
            $row\_[0][3] = orig\_race$\\
            $row\_[0][4] = orig\_sex$\\
            
        }
        $num\_unique\_vals = get\_unique(pred\_list)$\\
        \If{$num\_unique\_vals > 1$}{
            removal\_list.append($index$)\\
        }
        \If{num\_unique\_vals $==$ 0}{
            $raise EmptyList$\\
        }
    }
   
    removal\_list = set(removal\_list)\\
    total\_biased\_points = $len$ of removal\_list\\
    total\_dataset\_points = df\_test.shape[0]\\

    AWI = (total\_biased\_points / total\_dataset\_points)\\
    return AWI\\

}
\end{algorithm}

\subsection{Experimental Design}
Here we describe the process we took to prepare our data for our experiments. Our study uses the 2020 HMDA data for the training and testing set and Logistic Regression (LSR) as the classification model. For each experiment, the dataset is split using 5-fold cross-validation (train - 70\%, test- 30\%). This step is repeated 10 times with random seed and then the median is reported. The feature columns that have at most 10\% of the values as missing or not applicable are kept, but any rows containing said values that are missing or not applicable are removed. Additionally, non-numerical features are converted to numerical (e.g female: 0, male: 1, joint: 2) values. It is important to note that any data points that do not contain white, black or join as race are removed. Finally, all feature values are normalized between 0 and 1.

\begin{table*}
\centering
\caption{Results before and after DualFair. AWI and False Alarm the lower the better. Accuracy, Precision, Recall, and F1 Score the higher the better. “Green” cells show improvement and “Red” cells show damage. “Gray” cells show no change. All values are rounded. AWI is multiplied by 10 to accurately show the change.}
\begin{adjustbox}{width= 18.3cm, height = 3.75cm}
\begin{tabular}{|l|l|c|c|c|c|c|c|c|c|c|c|c|c|c} 
\cline{1-14}
State                                                                      & \# of rows                  & \multicolumn{2}{c|}{AWI (-)}                                                      & \multicolumn{2}{c|}{Accuracy (+)}                                                  & \multicolumn{2}{c|}{Precision (+)}                                                 & \multicolumn{2}{c|}{Recall (+)}                                                    & \multicolumn{2}{c|}{False Alarm (-)}                                               & \multicolumn{2}{c|}{F1 Score (+)}                                                  &                       \\ 
\cline{1-14}
\multicolumn{1}{|c|}{}                                                     & \multicolumn{1}{c|}{}       & Before               & After                                                      & Before                & After                                                      & Before                & After                                                      & Before                & After                                                      & Before                & After                                                      & Before                & After                                                      &                       \\ 
\hhline{|--------------~}
\multirow{2}{*}{\begin{tabular}[c]{@{}l@{}}Nationwide \\HMDA\end{tabular}} & \multirow{2}{*}{12 million} & \multirow{2}{*}{0.2} & {\cellcolor[rgb]{0.659,0.816,0.553}}                       & \multirow{2}{*}{0.98} & {\cellcolor[rgb]{0.976,0.514,0.514}}                       & \multirow{2}{*}{0.99} & {\cellcolor[rgb]{0.659,0.816,0.553}}                       & \multirow{2}{*}{0.98} & {\cellcolor[rgb]{0.976,0.514,0.514}}                       & \multirow{2}{*}{0.03} & {\cellcolor[rgb]{0.659,0.816,0.553}}                       & \multirow{2}{*}{0.98} & {\cellcolor[rgb]{0.976,0.514,0.514}}                       &                       \\
                                                                           &                             &                      & \multirow{-2}{*}{{\cellcolor[rgb]{0.659,0.816,0.553}}0.06} &                       & \multirow{-2}{*}{{\cellcolor[rgb]{0.976,0.514,0.514}}0.97} &                       & \multirow{-2}{*}{{\cellcolor[rgb]{0.659,0.816,0.553}}1.00} &                       & \multirow{-2}{*}{{\cellcolor[rgb]{0.976,0.514,0.514}}0.94} &                       & \multirow{-2}{*}{{\cellcolor[rgb]{0.659,0.816,0.553}}0.00} &                       & \multirow{-2}{*}{{\cellcolor[rgb]{0.976,0.514,0.514}}0.97} &                       \\ 
\hhline{|--------------~}
\begin{tabular}[c]{@{}l@{}}CA \\(Large)\end{tabular}                       & 3,423,897                   & 0.17                 & {\cellcolor[rgb]{0.659,0.816,0.553}}0.06                   & 0.97                  & {\cellcolor[rgb]{0.753,0.753,0.753}}0.97                   & 0.99                  & {\cellcolor[rgb]{0.659,0.816,0.553}}1.00                   & 0.97                  & {\cellcolor[rgb]{0.976,0.514,0.514}}0.94                   & 0.03                  & {\cellcolor[rgb]{0.659,0.816,0.553}}0.00                   & 0.98                  & {\cellcolor[rgb]{0.976,0.514,0.514}}0.97                   &                       \\ 
\hhline{|--------------~}
\begin{tabular}[c]{@{}l@{}}TX \\(Large)\end{tabular}                       & 1,964,077                   & 0.13                 & {\cellcolor[rgb]{0.659,0.816,0.553}}0.02                   & 0.98                  & {\cellcolor[rgb]{0.753,0.753,0.753}}0.98                   & 0.99                  & {\cellcolor[rgb]{0.659,0.816,0.553}}1.00                   & 0.98                  & {\cellcolor[rgb]{0.976,0.514,0.514}}0.97                   & 0.02                  & {\cellcolor[rgb]{0.659,0.816,0.553}}0.00                   & 0.99                  & {\cellcolor[rgb]{0.976,0.514,0.514}}0.98                   &                       \\ 
\hhline{|--------------~}
\begin{tabular}[c]{@{}l@{}}IL\\(Medium)\end{tabular}                       & 864,270                     & 0.06                 & {\cellcolor[rgb]{0.976,0.514,0.514}}0.11                   & 0.98                  & {\cellcolor[rgb]{0.753,0.753,0.753}}0.98                   & 0.99                  & {\cellcolor[rgb]{0.659,0.816,0.553}}1.00                   & 0.99                  & {\cellcolor[rgb]{0.976,0.514,0.514}}0.97                   & 0.04                  & {\cellcolor[rgb]{0.659,0.816,0.553}}0.00                   & 0.99                  & {\cellcolor[rgb]{0.976,0.514,0.514}}0.98                   &                       \\ 
\hhline{|--------------~}
\begin{tabular}[c]{@{}l@{}}WA~\\(Medium)\end{tabular}                      & 823,323                     & 0.18                 & {\cellcolor[rgb]{0.976,0.514,0.514}}0.26                   & 0.98                  & {\cellcolor[rgb]{0.976,0.514,0.514}}0.97                   & 0.99                  & {\cellcolor[rgb]{0.659,0.816,0.553}}1.00                   & 0.98                  & {\cellcolor[rgb]{0.976,0.514,0.514}}0.95                   & 0.05                  & {\cellcolor[rgb]{0.659,0.816,0.553}}0.00                   & 0.99                  & {\cellcolor[rgb]{0.976,0.514,0.514}}0.97                   &                       \\ 
\hhline{|--------------~}
\begin{tabular}[c]{@{}l@{}}NV \\(Small)\end{tabular}                       & 316,969                     & 0.13                 & {\cellcolor[rgb]{0.659,0.816,0.553}}0.09                   & 0.97                  & {\cellcolor[rgb]{0.659,0.816,0.553}}0.98                   & 0.98                  & {\cellcolor[rgb]{0.659,0.816,0.553}}1.00                   & 0.98                  & {\cellcolor[rgb]{0.976,0.514,0.514}}0.97                   & 0.05                  & {\cellcolor[rgb]{0.659,0.816,0.553}}0.00                   & 0.99                  & {\cellcolor[rgb]{0.976,0.514,0.514}}0.98                   &                       \\ 
\hhline{|--------------~}
\begin{tabular}[c]{@{}l@{}}CT~\\(Small)\end{tabular}                       & 231,251                     & 0.24                 & {\cellcolor[rgb]{0.659,0.816,0.553}}0.05                   & 0.98                  & {\cellcolor[rgb]{0.753,0.753,0.753}}0.98                   & 0.99                  & {\cellcolor[rgb]{0.659,0.816,0.553}}1.00                   & 0.98                  & {\cellcolor[rgb]{0.976,0.514,0.514}}0.96                   & 0.04                  & {\cellcolor[rgb]{0.659,0.816,0.553}}0.00                   & 0.99                  & {\cellcolor[rgb]{0.976,0.514,0.514}}0.98                   & \multicolumn{1}{l}{}  \\ 
\cline{1-14}
\multicolumn{1}{l}{}                                                       & \multicolumn{1}{l}{}        & \multicolumn{1}{l}{} & \multicolumn{1}{l}{}                                       & \multicolumn{1}{l}{}  & \multicolumn{1}{l}{}                                       & \multicolumn{1}{l}{}  & \multicolumn{1}{l}{}                                       & \multicolumn{1}{l}{}  & \multicolumn{1}{l}{}                                       & \multicolumn{1}{l}{}  & \multicolumn{1}{l}{}                                       & \multicolumn{1}{l}{}  & \multicolumn{1}{l}{}                                       & \multicolumn{1}{l}{} 
\end{tabular}
\end{adjustbox}
\end{table*}

\begin{table*}
\centering
\caption{Fairness and Performance Metrics }
\begin{adjustbox}{width=16.35cm, height= 3.0cm,center}
\begin{tabular}{|l|l|} 
\hline
\rowcolor[rgb]{0.82,0.82,0.82} Performance Metric                                                                                                                                                                      & Ideal Value           \\ 
\hline
\textbf{\textbf{Accuracy~}}~~=~  $\dfrac{(TP + TN)}{(TP + FP + TN + FN)}$                                                                                                                                                                                          & 1                     \\ 
\hline
\textbf{\textbf{Precision~}}~ =~TP/(TP+FP)                                                                                                                                                                             & 1                     \\ 
\hline
\textbf{False Alarm} =~ FP/N = FP/(FP+TN)                                                                                                                                                                              & 0                     \\ 
\hline
\textbf{Recall }=~TP/P = TP/(TP+FN)                                                                                                                                                                                    & 1                     \\ 
\hline
\textbf{F1 Score~}=  $\dfrac{2 * (Precision * Recall)}{(Precision + Recall)}$                                                                                                                                                                                                    & 1                     \\ 
\hline
\rowcolor[rgb]{0.82,0.82,0.82} Fairness Metric                                                                                                                                                                         & Ideal Value           \\ 
\hline
\begin{tabular}[c]{@{}l@{}}\textbf{Alternate World Index (AWI):} the normalized number of biased points in a dataset,\\where biased points are classified by a certain models consistency in predictions.\end{tabular} & 0                     \\ 
\hline
\multicolumn{1}{l}{}                                                                                                                                                                                                   & \multicolumn{1}{l}{} 
\end{tabular}
\end{adjustbox}
\end{table*}

Now, we will describe how we get results for each of our experiments. In DualFair, the training and testing data are both repaired during the bias mitigation pipeline. We evaluate AWI fairness metrics before and after the bias mitigation process for comparison. To do this, we first train a classification model on the training data (i.e either before or after DualFair) and then measure its fairness and accuracy on testing data. Our accuracy is measured in terms of recall, \emph{false alarm}, \emph{accuracy}, \emph{F1 score}. Fairness is measured in \emph{AWI}. \emph{Recall}, \emph{accuracy}, and \emph{F1 score} are better at larger values (i.e closer to 1). \emph{False alarm} and \emph{AWI} are better at smaller values (i.e closer to 0) 

In this work, we perform experiments using DualFair on 2020 state-level data from 2 small states (<150,000 rows of data), 2 medium states (>150,000 to <250,000 rows of data), and 2 large states (>250,000 rows of data). We also perform an experiment using DualFair on 2018-2020 nationwide HMDA (>15,000,000 rows of data). For this experiment, we randomly sample 5 million rows from each year and then apply DualFair. In terms of finding our 2 small, medium, and large states, we group according to category and randomly sample 2 states from each group.




\section{Results}
We structure our results around 5 central research questions. 

\subsection{RQ1: How well does DualFair create a MSPSO fair-loan predictor?}

\textbf{RQ1} explores the performance of our pipeline in debiasing mortgage data and creating a fair-loan-predictor. It is reasonable to believe that a fair-loan predictor should do two things proficiently: prediction and fairness. Accordingly, we test DualFair for both performance metrics (e.g accuracy, recall, precision, and F1 score) and fairness metrics (e.g AWI).  A summary of all the metrics we use can be found in Table 2. 

In Table 3, we give performance and fairness before and after the DualFair pipeline. We run 7 different trials from a range of small, medium, large states, and nationwide of HMDA. Rows 2, 3, 4, 5, 6, 7, and 8 summarize the results of DualFair on the trials. Measured in terms of AWI, DualFair is successful in increasing fairness for 5 of 7 states varying in data size. It is important to note that AWI is multiplied by 10 to more accurately represent the difference between the before and after of the bias mitigation pipeline. In addition, DualFair benefits precision and false alarm on all occasions while damaging recall and F1 score only slightly.  

Thus the answer to \textbf{RQ1} is “DualFair establishes a MSPSO fair-loan predictor that achieves both high levels of accurate prediction and fairness.” This means that DualFair can mitigate bias within MSPSO data while maintaining high levels of fairness. This is one of the biggest achievements of our work and is pivotal towards real-world applicability.

\subsection{RQ2: How does DualFair’s fair-loan-predictor compare to the state-of-the-art?}

\textbf{RQ2} serves to compare DualFair to other state-of-the-art models that have been evaluated on nationwide HMDA. For our study, the only model remotely similar to DualFair is used in Gill et al. 

Gill et al. uses FPR (False Postive Rate) to measure fairness. Even Gill et al., however, can not be used as a comparison to the state-of-the-art because of prediction on a different label and varied metric usage. 

Therefore, for RQ2, we point out the fact that our accuracy was higher than Gill et al. in terms of predicting  our respective label. We believe our removal of the accuracy-fairness tradeoff leads to this result. Nevertheless, there is no viable model for comparison against DualFair that can be reasonably be applied to HMDA. 

Further, it can be concluded that the solution to \textbf{RQ2} is “DualFair generally performs at the level of other state-of-the-art models but can not be compared directly to another bias mitigating model in the fairness domain due to the lack of work.” We hope that in the future we might be able to compare our DualFair in the MSPSO domain.

\subsection{RQ3: Does DualFair eliminate the accuracy-fairness tradeoff?}

\textbf{RQ3} seeks to consider if DualFair effectively removed the accuracy-fairness tradeoff. That is, we hope to explore if our pipeline resulted in consistent accuracy while increasing fairness simultaneously. 

In all our testing, including our 7 rigorous trials in table 3, we found that our accuracy remained consistent prior to debiasing. It also remain stagnate after debiasing, albeit an increase in fairness metrics. We hypothesize that the reason for this derives from our pipeline which "repairs" both training and testing data. 

Hence, the answer to \textbf{RQ3} is “Yes, DualFair simultaneously removes the accuracy-fairness tradeoff while achieving individual fairness”

\subsection{RQ4: Is DualFair capable of capturing all MSPSO in HMDA?}
To answer \textbf{RQ4}, we established forecasted models with increasing MSPSO data for HMDA using DualFair. The literature has suggested that ASPSO, or all sensitive parameters and sensitive options, may not be both computationally and logically feasible due to data division into very small, unmanageable regions.

Our experience shows us that it may be feasible depending upon an adequate pipeline and particular data conditions. Our simplified models tell us that it is possible for DualFair to scale to ASPSO. We determined that although computational expense compounds, our code is still adequate at scaling with the demand of more sensitive parameters or options.

One limitation of this conclusion is adequate data must be provided. That is, data which contains at least two individuals from all ASPSO combinations must be present. In addition, each group must contain at least a rejected individual (y = 0) and an accepted individual (y = 1) to generate oversamples using SMOTE. Thus, there becomes a problem with data fragmentation into small sub-datasets when large MSPSO data may not be readily available.

Overall, it can be concluded that DualFair has the capacity to scale to all of HMDA, given its large dataset size that can provide adequate data even to large division of datasets.

\section{Future Works}

DualFair, albeit comprehensive, has room for future improvements. We propose a variety of possibilities to enhance working with MSPSO  and mitigating bias. The following are future directions and limitations for researchers to consider: 

\begin{itemize}
    \item 
     \textbf{Direction 1:} Use of DualFair within other AI-related domains (e.g healthcare, job applications, and car insurance). 
    \item
    \textbf{Direction 2:}  Performing a widespread analysis of bias mitigating methods (including DualFair) with AWI as a fairness metric for comparison.
    \item
    \textbf{Direction 3:} Applying DualFair or another bias mitigating method to ML regression models in various domains. 
    \item
    \textbf{Limitation 1:} Instability of DualFair on low amounts of MSPSO data.
    \item
    \textbf{Limitation 2:} DualFair, although it removes the accuracy-fairness tradeoff, needs large quantities of data to achieve high-performance metrics.
    
\end{itemize}

We provide our GitHub repository below to aid practitioners and researchers in enhancing the domain of AI fairness and/or adopting DualFair for their own purposes. The repository contains the source code for DualFair and AWI as well as their application on HMDA data  (\url{https://github.com/ariba-k/fair-loan-predictor}).

\section{Conclusion}
This paper has tested the DualFair process, which removes label bias and selection bias within the working data (\emph{pre-processing}). As shown above, we have shown that DualFair can be applied to the HMDA dataset to create a high-performing fair ML classifier in the mortgage lending domain. Unlike other ML fairness pipelines, DualFair is capable of such results even if the data contains non-binary \emph{sensitive parameters} and \emph{sensitive options}, such as in the case of HMDA. We have shown that DualFair is a comprehensive bias mitigation tool that can also be generalized. In summary, we have done the following in this work: 

\begin{enumerate}
    
\item
Created a novel bias mitigating method called DualFair for data with MSPSO.

\item
Developed a new fairness metric (\emph{AWI}) that can be applied to MSPSO data. 

\item
Established a fair machine learning classifier in the mortgage lending domain.

\end{enumerate}

We aim for our work to cause the creation of fair ML models in other domains of work and solve the dilemma of linking research with real-world deployment. We hope our work is adopted by mortgage lending organizations wanting to implement a state-of-the-art non-discriminatory model for loan prediction and comply with the ECOA. 

\bibliographystyle{unsrt}
\bibliography{Ethics.bib}

\end{document}